\documentclass{article}
\usepackage{spconf,amsmath,graphicx}
\usepackage{multirow}

\title{Trustera: A Live Conversation Redaction System}
\name{\parbox{\linewidth}{\centering Evandro Gouvêa, Ali Dadgar, Shahab Jalalvand, Rathi Chengalvarayan\sthanks{The author performed the work while at Interactions LLC.},  Badrinath Jayakumar, \\
Ryan Price$^*$, Nicholas Ruiz, Jennifer McGovern, Srinivas Bangalore, Ben Stern}}

\address{Interactions LLC, New Providence, New Jersey, USA}

\begin{document}
%
\maketitle

\begin{abstract}
We introduce Trustera\footnote{https://www.interactions.com/products/trustera/}, the first functional system that redacts personally identifiable information (PII) in real-time spoken conversations to remove agents' need to hear sensitive information while preserving the naturalness of live customer-agent conversations. As opposed to post-call redaction, audio masking starts as soon as the customer begins speaking to a PII entity. This significantly reduces the risk of PII being intercepted or stored in insecure data storage. Trustera’s architecture consists of a pipeline of automatic speech recognition, natural language understanding, and a live audio redactor module. The system’s goal is three-fold: redact entities that are PII, mask the audio that goes to the agent, and at the same time capture the entity, so that the captured PII can be used for a payment transaction or caller identification. Trustera is currently being used by thousands of agents to secure customers’ sensitive information.
\end{abstract}


\section{Introduction}
\label{sec:intro}

Gartner Research projects \cite{gartner2020} that the customer experience workforce that work from home (WFH) will increase from 5\% in 2017 to 35\% by 2023. 
This presents an increased risk of breaches of sensitive customer data.
An IBM work from home study in 2020 reported that 42\% of United States-based respondents who were newly WFH worked with PII in their job. 
More than half had not been provided with updated guidelines on how to handle PII while working from home, despite being required to handle PII to support customer service requests\footnote{https://newsroom.ibm.com/2020-06-22-IBM-Security-Study-Finds-Employees-New-to-Working-from-Home-Pose-Security-Risk}. 
%
In addition to PCI-DSS violations, payment information falls under the scope of GDPR, and violations can cost businesses up to €20 million or 4\% of the annual worldwide turnover of the preceding financial year\footnote{https://gdpr.eu/fines/}.

In order to safeguard customer PII, we present Trustera, a real-time dialog mediation system that intercepts and redacts sensitive personal information so call center agents can complete natural conversations with customers without needing to hear PII such as payment card information (PCI) and social security numbers (SSNs).
To prevent agents from hearing customer PII, Trustera mutes the audio sent to the agent as soon as it detects the beginning of PCI, such as payment card numbers and the security code on the back of the card (CVV).
Trustera is currently being used by thousands of agents to secure customers’ sensitive information.
Its architecture consists of a pipeline of online automatic speech recognition (ASR), natural language understanding (NLU) and a live audio redactor (LAR) module. 

To the best of our knowledge, Trustera is the first AI-driven audio redaction service that redacts PII in real-time and protects the entire call. 
Other previously proposed solutions use DTMF suppression or Conversational AI chatbots to protect PII, but they require agents to manually turn on the service in order to protect a portion of a call.
While the concept of redaction is not new, most redaction experiments \cite{stubbs2015automated,neamatullah2008automated} cover written text, which do not contain noisy input induced by ASR errors. 
Some commercial companies\footnote{https://docs.aws.amazon.com/transcribe/latest/dg/pii-redaction.html} provide an ASR redaction service for streaming ASR transcripts, but it does not mask PII in the audio, so the call is not protected in real-time.
Trustera identifies and redacts PII entities in the audio as soon as the entity is uttered, minimizing the exposure of PII to agents and to downstream call recordings.
Offline audio de-identification \cite{cohn2019audio} is the closest to our work; however, it redacts PII on pre-recorded audio as opposed to live conversations and does not have latency constraints.

Another related research area is spoken entity detection. \cite{ghannay2018end} experiment with cascading ASR followed by NLU module, and compares it to an integrated end-to-end approach.
Currently Trustera uses the cascaded pipeline approach in order to minimize the CPU load, since call center servers usually lack GPUs. 
End-to-end approaches usually require GPUs or they consume a lot of CPU resource, which is cost-prohibitive for call centers.
Notable works of cascading ASR and NLU include \cite{hatmi2013named} and \cite{sudoh2006incorporating}.

\section{System Architecture}
\label{sec:arch}
 
\begin{figure}[htb]
\centering
\includegraphics[trim=0cm 0cm 0cm 0cm, clip=true, width=\columnwidth]{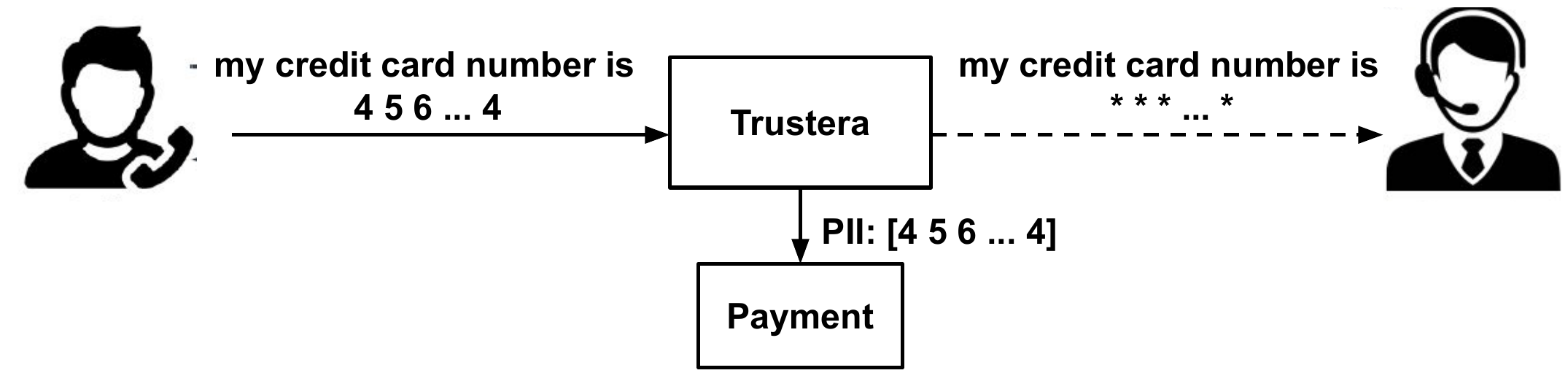}
\caption{Real-time PII redaction.}
\label{fig:exp}
\end{figure}

Trustera's goal is to (1) identify PII entities, (2) mask their mention in the audio and transcripts, and (3) capture the entity values for secure transaction processing.
Figure \ref{fig:exp} demonstrates secure credit card number redaction by Trustera. We describe an early lab version of our solution in the paragraphs below.

Incoming stereo audio is decoded with two ASR decoders for the agent and caller. 
Live audio redactor (LAR) monitors the partial hypotheses and it triggers the NLU module as soon as a digit is detected in the hypothesis. 
NLU predicts the type of the entity and it informs LAR if the coming entity needs to be redacted.
Then, LAR masks the audio till the end of entity is decided.
The canonical value of the entity is extracted and sent to the payment system to complete the transaction.

\subsection{ASR module}
\label{ssec:asr}

The ASR module is a finite state machine (FSM) consisting of a hybrid DNN-HMM BiLSTM acoustic model \cite{hinton2012deep}, a 4-gram language model \cite{katz1987estimation} and a pronunciation dictionary.

The acoustic model consists of bidirectional long short-term memory (BiLSTM) architecture with a projection layer trained in PyTorch to predict tied context-dependent triphone HMM states and optimized with cross-entropy with stochastic gradient descent. 
There are 108 phonemes and approximately 42,369 tied context-dependent HMM states modeled in the output layer. 
The network is trained on 11 million utterances sampled at 8kHz with a mu-law compression, about 40,000 hours of anonymized and hand-transcribed utterances, collected from a variety of customer care applications. 
The input features used for training the acoustic model are represented with 20ms frames of 81-dimensional log-spectrum features computed every 10ms, with no frame stacking.
The network has four layers each with 600 memory cells and 300 projection units with a total number of about 41M model parameters. 
8-bit quantization is applied to the acoustic model weight matrices to speed up the math-intensive operations of LSTM inference. Post training quantization is done using dynamic range mapping without calibration and results in a negligible loss in accuracy.

The language model (LM) is an interpolation of three 4-gram Katz back-off models \cite{katz1987estimation}: (1) trained on Fisher/ Switchboard \cite{cieri2004fisher} LM training data ($\approx$20M words); (2) trained on in-domain human transcribed data of ($\approx$7M words) and (3) trained on $\approx$10,000 calls automatically transcribed using an offline/accurate end-to-end pre-trained ASR \cite{kuchaiev2019nemo} ($\approx$10M words). 
The interpolation weights are tuned to minimize the perplexity of the resulting LM on a separate validation set.
The vocabulary is limited to the top 35K most frequent words with hand-crafted pronunciation for common words and grapheme-to-phoneme generated pronunciation for the rest of the vocabulary.

The stereo calls are decoded using two independent FSM decoders \cite{mohri2008speech}.
All the ASR decoder parameters are tuned to reduce the computation costs and increase accuracy. 
However, the trade-off between cost and the increased WER still favors CPUs.
Since the redaction must happen in real-time, the audio stream is decoded as soon as it reaches Trustera, and the partial ASR hypotheses are passed to the NLU and LAR modules for PII redaction. 

\subsection{NLU module}
\label{ssec:nlu}


The NLU model is a logistic regression classifier with a  liblinear solver and l2-norm penalty optimizer, implemented using sklearn \cite{pedregosa2011scikit}. 
The model is trained to detect the type of entity as soon as a digit word is recognized by the ASR decoder.
The feature vector is formed by the recognized digit followed by 20 previous tokens in the context. 
The context tokens are from both caller and agent channels merged.
We use 3-gram features to prepare the feature vector and concatenate a dialog state vector with binary features representing conversation time, previously detected entities, and transaction history.

We augment the training set using uncertainty-aware self-training \cite{mukherjee2020uncertainty}. 
To do so, a teacher model is trained using the available supervised data and then applied to several batches of unlabeled data.
Then, a combination of easy and hard samples are used to augment the training set. 

\subsection{LAR module}

LAR monitors the partial hypotheses generated by the ASR decoders from agent and caller channels. 
As soon as a digit is recognized in the partial hypothesis, LAR triggers the NLU module to predict the type of that entity.
If the predicted entity is a sensitive one to be redacted, then LAR masks the audio using a beep sound. 
Audio masking continues until the end of the entity is decided.
The end of the entity is a decided if the caller utters more than two non-digit words or silence of more than 3 seconds happens.
When boundary and type of the entity are detected, then LAR extracts the value of that entity and send it to the payment system.
Word to digit normalization is done by the LAR module based on well-defined grammars.
We avoid using DNN-based NLP models to normalize the entities simply for the sake of reducing CPU usage.

\subsection{Voice activity detection (VAD)}
Voice activity detection in Trustera is a two-stage process, where frames are marked as being speech or silence based on an energy-adaptive VAD method \cite{eadapt1996}, and the classification is then smoothed to determine the actual speech endpoints. Only frames classified as being speech are then sent to the decoder engine. Each channel decoder pipeline has its own VAD and end-pointer. We explored the alternative of detecting the voice activity before the audio is sent to each engine. Even though the idea of detecting the active channel and only decoding audio from that channel seems appealing, in practice it did not reduce CPU usage, and it reduced word accuracy, since speakers often overlap when speaking. Sending audio to only one channel at a time prevented the engine from fully recognizing what was spoken.

\subsection{Architecture considerations}
\label{ssec:arch}
Trustera is optimized for real-time processing with minimal CPU usage, so that audio redaction is triggered as close to the actual speaking time and at the same time maximum number of concurrent calls can be handled on a single CPU.
This is especially important for payment card security codes (\emph{e.g.}, CVV), which are only 3-4 digits in length. 
While advanced acoustic models (AMs) and end-to-end ASR models would increase transcription and entity classification accuracy, the large look-ahead requirements of the sequence models mean that the PII would be leaked before the audio frames were decoded.
This is because (1) end-to-end models are typically built on neural network layers with mechanisms that perform better with large input contexts like self-attention which need future context; (2) they use aggressive down-sampling to compress the length of the input sequence which reduces the granularity of any alignment between outputs and inputs;  and (3) they are trained with popular algorithms like CTC \cite{graves2006connectionist} or Transducer \cite{graves2012sequence} loss which learns to align the audio input sequence and label output sequence automatically by marginalizing over all possible valid alignments, without a strict requirement that the alignment corresponds to the physical occurrence of speech. 

\section{Experiments}

In this section, we report the relative results with regard to the best possible systems available as open source. For ASR, we show the WER performance with regard to an offline pretrained end-to-end ASR \cite{kuchaiev2019nemo}. For NLU, we report the performance with regard to an offline  BiLSTM-CRF classifier \cite{pressel2018baseline}.

\subsection{Data}

Table \ref{tab:call-stats} shows the statistics of the data set that is used to build the NLU modules.
The training set consists of 11K real calls between agents and customers. 
The PII were annotated by our internal labeling lab.
We annotated the type of PII, the actual words that were spoken when the customer was uttering the PII, and a canonical value representing the PII. 
The canonical value reflects the intended PII value written in a standard format (e.g. ``\emph{expiration date is February no no January twenty twenty five}'' $\rightarrow$ ``\emph{01/25}'').
The ASR models and the training data are described in Section \ref{ssec:asr}.

\begin{table}[t]
\centering
\footnotesize
\begin{tabular}{|l|r|r||l|r|r|}
\hline
Call stats & 	Train & Test & Entity stats & Train & Test\\
\hline
Calls & 11K & 359 & \textsc{routing} & 1234 & 129 \\
Hours & 1.4K & 51.5 & \textsc{bankacc} & 1126 & 124 \\
Words & 10.6M & 390K & \textsc{ccnum} & 8544 & 342 \\
Agent words & 6.9M & 245K & \textsc{cvv} & 5802 & 266 \\
Caller words & 3.7M & 144K & \textsc{expdate} & 8076 & 277 \\
Avg. call & \multirow{2}{*}{7.4} & \multirow{2}{*}{8.6}& \textsc{zip} & 7658 & 266 \\
duration (min) & & & \textsc{Others}  & 67K & 2.9K \\

\hline
\end{tabular}
\caption{Data statistics for the NLU module.} 
\label{tab:call-stats}
\vspace{-1em}
\end{table}

\subsection{Results}

We first evaluate Trustera's ASR performance and then the ASR+NLU which shows the performance of the whole system for redacting the PII entities in real-time.
All the inference experiments are conducted on CPUs. 
We run 8k (or 16k for the end-to-end ASR) samples/second to 30 jobs on servers with 12 core Intel(R) Xeon(R) CPU E5-2680 v4 @ 2.40GHz with multi-threading turned off. 
Each server had 130GB of memory, but actual memory usage was lower.

\subsubsection{ASR performance}

\begin{table*}[ht]
    \centering
    \resizebox{1.0\linewidth}{!}{%
    \begin{tabular}{|l|r|r|r|r||r|r|r|r|r|r|}
    \hline
    \multicolumn{1}{|c|}{ASR} &  \multicolumn{1}{c|}{CPUvsAudio$\downarrow$} & \multicolumn{3}{c||}{Relative Overall  WER$\downarrow$} & \multicolumn{6}{c|}{Relative WER/SER of Entities $\downarrow$}   \\
    \hline
    &  & Agent & Caller & Conv & ABAROUTING & BANKACCOUNT & CCNUM & CVV & EXPDATE & ZIP  \\ \hline
    E2E-pretrained-offline & -- & -- & -- & -- & --/-- & --/-- & --/-- & --/-- & --/-- & -- \\
    E2E-tuned-offline & 0.0 & -29.0 & -10.3 & -20.5 & +20.4/+7.6 & -9.1/-12.3 & +6.4/+16.2 & +3.1/+26.4 & +10.7/+2.3 & -40.5/-44.5 \\
    E2E-tuned-online & -78.7 & +23.8 & +12.3 & +18.4 & +120.4/+58.3 & +230.3/+12.3 & +71.8/+21.4 & +8.6/+59.3 & +13.1/+18.1 & +6.0/+22.7 \\
    Trustera-online & -53.2 & -21.5 & -8.2 & -7.8 & -100.0/-100.0 & -51.5/-58.5 & -71.8/-35.9 & -33.3/-11.6 & -21.4/-9.0 & -45.2/-41.2 \\ \hline
    \multicolumn{1}{|c|}{NLU} &  \multicolumn{1}{c|}{CPUvsAudio$\downarrow$} & \multicolumn{3}{c||}{Relative Overall  WER$\downarrow$} & \multicolumn{6}{c|}{Relative Precision/Recall drop of Entities $\uparrow$}  \\ \hline    
    Trustera+BiLSTM-CRF-offline & -- & -21.5 & -8.2 & -7.8 & --/-- & --/-- & --/-- & --/-- & --/-- & \\
    Trustera+Log-Reg-online & -53.2 & -21.5 & -8.2 & -7.8 & -22.8/-10.3 & -16.5/-9.3 & -24.3/-11.6 & -8.0/+3.7 & -10.0/-0.8 & -2.6/-5.2   \\ \hline
    \end{tabular}%
}
    \caption{Trustera's Performance in terms of real-time factor (CPUvsAudio), word error rate (WER) and sentence error rate (SER) and Precsion/Recall for detecting the boundary and type of the entities. $\downarrow$ implies the more negative the better. $\uparrow$ implies the less negative the better.}
    \label{tab:werser}
    \vspace{-1em}
\end{table*}

The ASR performance is measured by word error rate (WER) for the overall conversation and for the entity segments. 
The CPU usage is reported as CPUvsAudio, often called the real-time factor, and it is computed as the ratio between the CPU time, as reported by the operating system, and the audio duration. 

As the baseline ASR system, we use a publicly available off-the-shelf end-to-end ASR system from  Nvidia-Nemo  \cite{kuchaiev2019nemo}.
This is a non-autoregressive convolutional connectionist temporal classification (CTC) model \cite{majumdar2021citrinet}.
We use a greedy search decoder without any additional language model to decode the audio samples.
In Table \ref{tab:werser}, we refer to this system as \textit{E2E-pretrained-offline}.
All the ASR results in table \ref{tab:werser} are computed relatively with regard to this baseline system.

In order to show the impact of the in-domain data, we fine-tune the end-to-end model using our training set in Table \ref{tab:call-stats}.
We continue the tuning step for 100 epochs and we call it \textit{E2E-tuned-offline}.
As it can be seen in Table \ref{tab:werser}, the real-time factor doesn't change, however the overall WER improves by a factor of 20.5\%. 
This improvement is larger on the agent side (29.0\%) and smaller on the caller side (10.3\%). 
Although the tuned model works well on the overall conversation, it fares poorly in recognizing the entity segments. 
Except for bank account and zip codes, where the WER of the tuned system reduces, all other entities got worse.  
We believe this happens due to the lack of enough entity samples in the tuning set. We had to remove numeric PIIs from the tuning set, as we are not allowed to keep intact PII in the servers running the E2E engine.

Then, we use the tuned end-to-end model and run the decoder in a an online mode  (\textit{E2E-tuned-online}).
We experience salient reduction in CPU usage by a factor of 78.7\%, but at the same time all WER and SER measures degrade dramatically compared to \textit{E2E-pretrained-offline}.  

Finally we evaluate Trustera in an online mode.
As mentioned in section \ref{ssec:asr}, we use a hybrid BiLSTM acoustic model along with a 4-gram language model for the ASR module. 
As it can be seen, Trustera's online ASR decoder reduces the CPU consumption by 53.2\%.
Moreover, it improves the WER by 21.5\%, 8.2\% and 7.8\% respectively for agent, caller and overall. 
It also outperforms the baseline system when recognizing all the entity segments. 
For example, all the routing numbers are recognized correctly with Trustera's ASR.
Also WER/SER of the credit card numbers improved substantially.
Unlike the end-to-end models, where the acoustic model is pretrained on generic and diverse (e.g. wideband audio, clean speech etc) data and then fine-tuned on the in-domain sets, Trustera's acoustic model is trained from scratch on in-domain data and other data that are acoustically very similar to the call center audio (e.g. voicemail). 
Moreover, Trustera takes advantage of an in-domain language model that has been trained on a large amount of digit sequences. 

Results show that although this early lab system uses older ASR technology compared to the state-of-the-art end-to-end system, its performance is efficient in terms of CPU usage and accuracy. 
Note that usually the end-to-end systems need GPU which is not available in the call-center servers. 

\subsubsection{NLU performance}

The second part of Table \ref{tab:werser} shows the precision/recall difference between an oracle offline NLU and our online model used in Trustera.
A prediction is counted as true positive if the begin, end and type of the entity is predicted correctly.
The same ASR system is used to generate the hypotheses for the NLU modules. 
The offline NLU takes advantage of fully stable hypotheses whereas the online NLU uses unstable (partial) hypotheses only.

As the baseline, we use an offline BiLSTM-CRF model trained on the 11k calls (in Table \ref{tab:call-stats}) and it is applied to the stable ASR hypotheses with full access to the left and right context of the dialogue (\textit{BiLSTM-CRF-offline}).

Trustera's NLU uses a logistic regression model, due to the architecture constraints, and it has access to the left context only (\textit{Log-Reg-online}).
This model is also trained on 11k calls, but applied to the unstable hypotheses in order to mimic the real-life application.
Considering CCNUM, as one of the most important entities, we observe that Trustera's NLU precision drops relatively by 24.3\% compared to BiLSTM-CRF.
We believe that drop is due to the BiLSTM-CRF model benefiting from backward information (i.e. observing a 16-digit credit card number).
Recall is a more relevant measure of redaction performance.
We observe that CCNUM recall drops by 11.6\%, relatively.
In the case of CVV, Trustera indeed outperforms BiLSTM-CRF by 3.7\%.

Our experiments show that there is a large gap between stable and unstable hypotheses when it comes to offline NLU versus online. 
Moreover, the right context in BiLSTM-CRF plays an important role in predicting the type of entity.

\subsection{Analysis}
Trustera may leak sensitive entities too.
The cause of the leak usually comes from ASR errors, word-to-digit normalization, customer hesitations and agent interruptions.
For example, the PII start detector will not fire if an entity's beginning digit is misrecognized as a non-digit token. 
Also the word-to-digit normalization may fail if the interpretation is ambiguous like, ``\emph{one hundred twenty} $\rightarrow$ \emph{120} or \emph{100-20}'' or  ``\emph{double oh seven}''.
 

\section{Conclusion}
Trustera, a real-time dialog mediation system that intercepts and redacts sensitive personal information, was presented. It is the first system that monitors and redacts the sensitive information from entire calls in real-time so that the agent cannot hear customer PII. Trustera redacts and captures payment information without interrupting the natural conversation, and sends the information to the payment system. Trustera significantly reduces the risk of sensitive information leaks, and is highly efficient in terms of CPU usage and latency.



\vfill\pagebreak

\bibliographystyle{IEEEbib}
\bibliography{main}

\end{document}